\documentclass[mlmain]{jmlr}

\usepackage{longtable}

\usepackage{booktabs}
\usepackage[load-configurations=version-1]{siunitx} 

\usepackage{tikz}           
\usepackage{ccaption}       

\theorembodyfont{\upshape}
\theoremheaderfont{\scshape}
\theorempostheader{:}
\theoremsep{\newline}

\title[V1 Neuronal Representations Improve Robustness]{Matching the Neuronal Representations of V1 is Necessary to Improve Robustness in CNNs with V1-like Front-ends}

\author{Ruxandra Barbulescu$^{*,1}$, Tiago Marques$^{*,2}$, Arlindo L. Oliveira$^{1,3}$\\
        \\
        $^*$ Joint first authors (equal contribution) \\
        $^1$ INESC-ID Lisboa, Portugal \\
        $^2$ Breast Unit, Champalimaud Clinical Center, Champalimaud Foundation, \\Lisboa, Portugal \\
        $^3$ IST Técnico Lisboa, Universidade de Lisboa, Portugal \\
        \\
        \texttt{ruxi@inesc-id.pt}  \hspace{20pt} \texttt{tiago.marques@research.fchampalimaud.org}
}

\begin{document}

\maketitle

\begin{abstract}
  While some convolutional neural networks (CNNs) have achieved great success in object recognition, they struggle to identify objects in images corrupted with different types of common noise patterns. Recently, it was shown that simulating computations in early visual areas at the front of CNNs leads to improvements in robustness to image corruptions. Here, we further explore this result and show that the neuronal representations that emerge from precisely matching the distribution of RF properties found in primate V1 is key for this improvement in robustness. We built two variants of a model with a front-end modeling the primate primary visual cortex (V1): one sampling RF properties uniformly and the other sampling from empirical biological distributions. The model with the biological sampling has a considerably higher robustness to image corruptions that the uniform variant (relative difference of 8.72\%). While similar neuronal sub-populations across the two variants have similar response properties and learn similar downstream weights, the impact on downstream processing is strikingly different. This result sheds light on the origin of the improvements in robustness observed in some biologically-inspired models, pointing to the need of precisely mimicking the neuronal representations found in the primate brain.

\end{abstract}

\section{Introduction}
\label{sec:intro}

Over the past decade, Convolutional Neural Networks (CNNs) have achieved great success in various computer vision tasks, namely object recognition \citep{Krizhevsky2012, simonyan_very_2015, he_deep_2016, szegedy_going_2015}. However, these models show a striking limitation in terms of robustness to image perturbations and out-of-domain generalization. CNNs struggle to recognize objects in images corrupted with different types of common perturbation patterns, such as random noise or weather effects, which humans excel at \citep{dodge_study_2017, geirhos_generalisation_2018, hendrycks_benchmarking_2019}.

Recently, there has been an increasing interest in incorporating circuit motifs found in biological brain circuits to address some of these limitations of CNNs \citep{malhotra_hiding_2020, dapello_simulating_2020, EVANS202296, baidya_combining_2021, cirincione_implementing_nodate}. One popular approach has been including model front-ends that simulate processing in the early stages of the visual system - retina, the lateral geniculate nucleus (LGN), and the primary visual cortex (V1). These front-ends usually consider Receptive Field (RF) structures that resemble those of neurons in V1. One popular implementation has been the use of a Gabor Filter Bank (GFB) \citep{jones_two-dimensional_1987} for the linear spatial kernels of a convolutional layer. Several of these models implement a GFB uniformly covering the RF parameter range, taking into consideration the resolution and scale of the input images. Dapello, Marques et al. went one step further and sampled the GFB parameters from known distributions of RF properties from primate V1 neuronal populations, observing that this led to an improvement in robustness against noise corruptions.

Here, we expand on this body of work to study the relationship between neuronal representations in low-level areas and robustness to image corruptions. We used the VOneNet model family \citep{dapello_simulating_2020} and created two variants of the VOneResNet18 model: one in which the GFB parameters were sampled from empirical distributions of primate V1 neuronal populations and another that samples the GFB parameters uniformly and independently within the same range. \textbf{This allows us to disambiguate the more general contribution of having a biological front-end with RFs inspired by those in primate V1 from precisely matching the same population-level statistics of those RF properties}. We make the following novel contributions:
\begin{enumerate}
    \item We reproduce the results from Dapello, Marques et al. in a different dataset (Tiny ImageNet, \cite{le_tiny_nodate}), observing a decrease in robustness to image corruptions when sampling the GFB parameters uniformly when compared to using the biological V1 neuronal distributions (relative decrease of 8.72\%).
    \item We observe that despite the very different distributions of neuronal RF properties and response properties between the two model variants, neuronal populations with similar RF properties and similar response properties across the two models contribute to downstream layers with similar weights.
    \item We show that the neuronal sub-populations in the V1 front-end that have the greatest impact in the downstream layers do not match between the two model variants, suggesting a likely reason for the observed differences in model robustness.
\end{enumerate}




\section{Methods}
\label{sec:methods}

Methods are explained in detail in the Appendix sections and a summary is presented below. 

We created two variants of the VOneResNet18 model (see Section \ref{sec:models_vonenets}, Figure \ref{fig:1} A and B): one sampling neuronal RF properties from biological distributions (Biological), and another sampling the properties uniformly and independently in the same range (Uniform). We trained (see Section \ref{sec:models_training}) four seeds of the two model variants as well as the standard ResNet18 model (see Section \ref{sec:models_resnet18}) on the Tiny ImageNet dataset (see Section \ref{sec:tiny}). We evaluated clean accuracy and robustness using the Tiny ImageNet-C dataset (see Section \ref{sec:corruptions}) and compared these with the baseline model (see Section \ref{sec:detailed_accuracies}).

\begin{figure} [b!]
\centering
\begin{tikzpicture}
\pgftext{\includegraphics[width=1\linewidth]{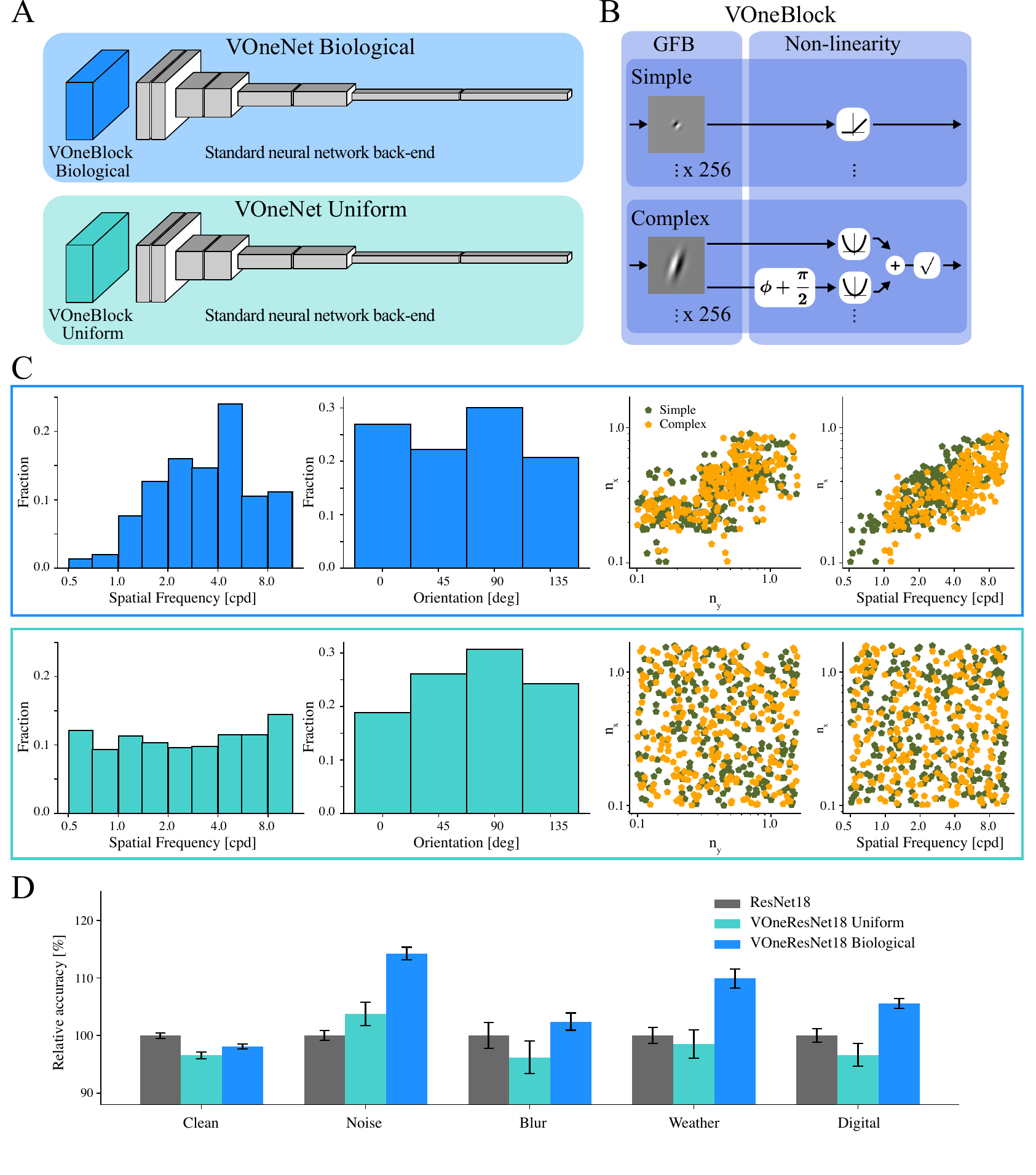}};
\end{tikzpicture}
\caption{(Continued on the following page.)}\label{fig:1}
\end{figure}

\begin{figure}[t!]
  \contcaption{\textbf{Matching the empirical distributions of RF properties of V1 neurons improves model robustness.} \textbf{A} The VOneNet model family contains a biologically-inspired model of V1, the VOneBlock, as the front-end to a standard CNN. We created two variants of the VOneBlock, one with RF parameters sampled from a uniform distribution (VOneBlock Uniform) and another with the parameters sampled from empirical distributions of primate V1 neurons (VOneBlock Biological). We used the ResNet18 as the standard CNN architecture. \textbf{B} The VOneBlock is a model of V1 with a GFB as a linear spatial RF, followed by a non-linear stage with simple- and complex-cell non-linearities. Here, we removed the stochasticity layer included in the original study. We considered 256 simple cell channels and 256 complex cell channels in both model variants. \textbf{C} Top, distribution of RF properties for one example seed of the VOneBlock Biological. Properties are sampled according to primate neurophysiological data. From left to right, spatial frequency, orientation, RF size as multiple of the Gabor wavelength ($n_x$ and $n_y$), and relationship between SF and RF size. Bottom, as above but for one seed of the VOneBlock Uniform. \textbf{D} Relative accuracy (normalized by the base model, ResNet18) for the two VOneNet variants for clean and corrupted images. Bars represent mean and error-bars represent s.e.m.. Note the higher robustness of the VOneResNet Biological when compared to the standard model and the model with uniform sampling ($n=4$). See \tableref{{tab:absolute_accuracies}} and Figure \ref{sup:detailed_accuracies} for absolute accuracies.}
\end{figure}

We studied the relationship between neuronal response properties (mean response activation and sparseness) (see Section \ref{sec:gfb_parameters}) and several neuronal RF properties: cell type (simple/complex), preferred spatial frequency (SF), and RF size (measured as multiples of the grating wavelength contained within one standard deviation of the Gaussian along the perpendicular axis, $n_x$) (see Section \ref{sec:rf_properties}). 

We grouped neurons into discrete neuronal sub-populations by their RF properties (cell type, SF, and $n_x$), and by their response properties (cell type, activation, and sparseness). For each sub-population, we calculated the mean absolute downstream weights after training (see Section \ref{sec:weights}). Finally, we estimated the Downstream Impact for each sub-population as the product of the number of neuronal channels, the mean response activation, and the mean absolute weights (see Section \ref{sec:impact}).

\section{Results}
\label{sec:results}




As intended, the variant with biological sampling has distributions of RF properties that resemble those found in empirical studies of primate V1 neuronal populations \citep{DeValois1982531, DeValois1982545, Ringach2002b, Schiller1976} while the variant with uniform sampling does not (Figure \ref{fig:1} C). While both VOneResNet18 variants have lower accuracies on clean images than the standard model, the model with biological sampling has substantially higher robustness against image corruptions for all corruption categories (average relative improvement of 7.85\%, Figure \ref{fig:1} D, Figure \ref{sup:detailed_accuracies},  and Table 1). The variant with uniform sampling, on the other hand, has considerably lower robustness when compared with the one with biological sampling (average relative decrease of 8.72\%).

\begin{figure} [h!]
\centering
\begin{tikzpicture}
\pgftext{\includegraphics[width=1\linewidth]{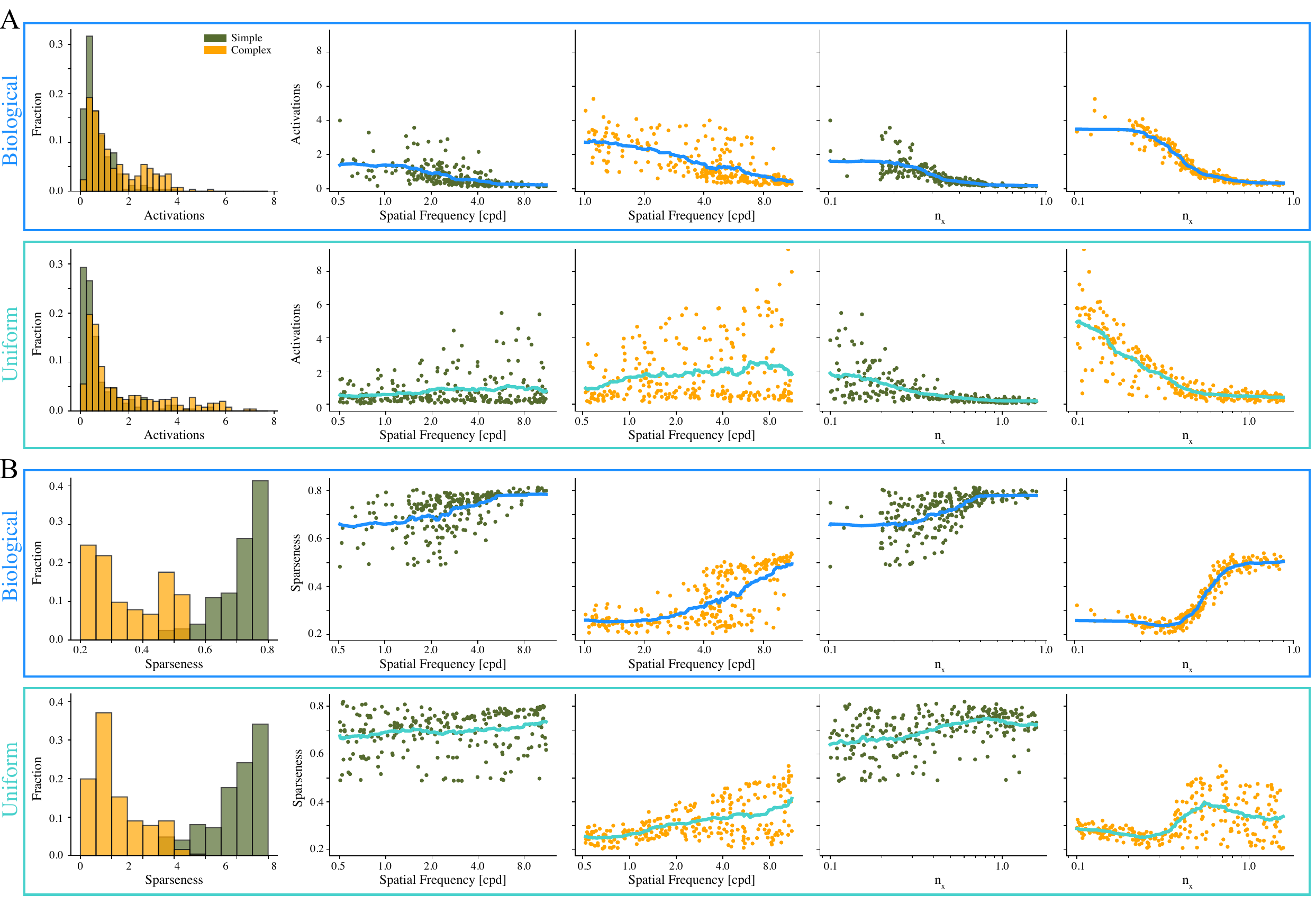}};
\end{tikzpicture}
\caption{\textbf{Neurons in the two VOneBlock variants show different response profiles.} \textbf{A} Top, mean response activation for neurons in one example seed of the VOneBlock Biological. From left to right, distribution of mean response activation for simple and complex cells, mean activation as a function of SF for simple cells, SF for complex cells, $n_x$ for simple cells, and $n_x$ for complex cells. Bottom, as above but for neurons in one seed of the VOneBlock Uniform. Note that while response activations decrease with increasing SF for the Biological variant, the same does not happen for the Uniform variant. \textbf{B} Same as \textbf{A} but for response sparseness. Thick line in plots represents the moving average (window size of 51). Trends were consistent in other model seeds.}\label{fig:2}
\end{figure}

To explore this difference in robustness between the two VOneResNet18 variants, we calculated two response properties of the VOneBlock neuronal channels (mean activation and sparseness) and compared these with the RF properties (cell type, SF, and $n_x$). In terms of sparseness, both variants have similar trends with simple cells showing considerably sparser responses, and sparseness increasing with SF and $n_x$ (Figure \ref{fig:2} B). Interestingly, there are notable differences in the relationships of the mean response activation and the GFB properties for the two variants (Figure \ref{fig:2} A). While activations decrease with increasing RF size in the two variants, for the Biological model there is also a decrease in response activation with increasing SF, which does not happen in the model with uniform sampling.

\begin{figure} [h!]
\centering
\begin{tikzpicture}
\pgftext{\includegraphics[width=0.8\linewidth]{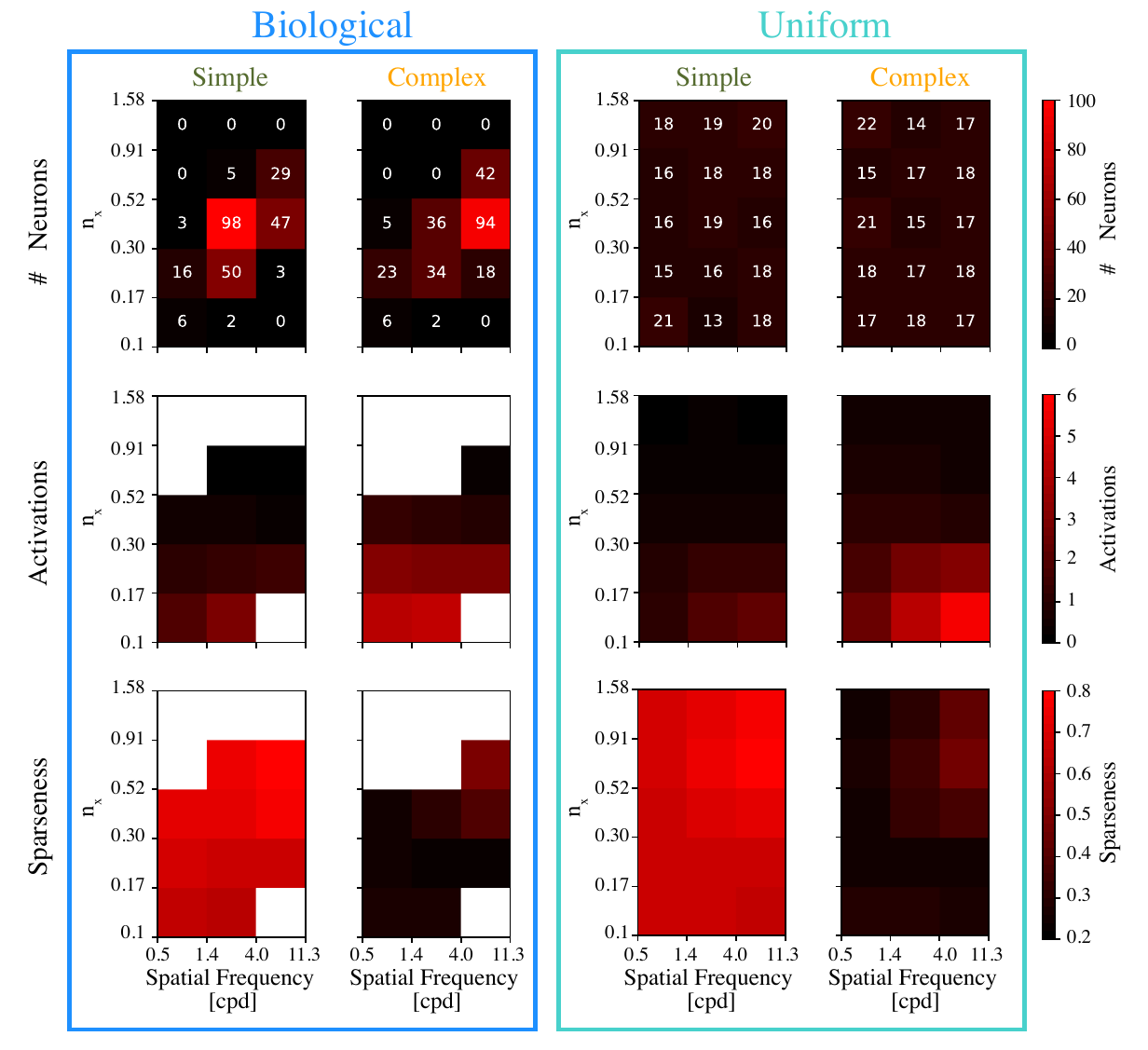}};
\end{tikzpicture}
\caption{\textbf{Effect of sampling from empirical distributions in model response activations.} Left, Biological Variant; right, Uniform Variant. Colormaps represent the average of all seeds ($n=4)$. Top row, distribution of number of neurons binned by SF and $n_x$ for simple and complex cells. Note that Biological Variant contains very few neurons with large $n_x$ and low SF, as well as neurons with low $n_x$ and high SF. Middle row, mean response activation for neurons according to the same binning. Neurons with low $n_x$ and high SF, which show the largest response activations, are substantially undersampled in the Biological Variant. Bottom row, same as above but for response sparseness. }\label{fig:3}
\end{figure}

Since the SF and $n_x$ are not sampled independently in the Biological variant, we grouped (binned) the neuronal channels in sub-populations according to the combination of their RF properties (cell type, SF, and $n_x$, Figure \ref{fig:3}). The model with biological sampling has a non-uniform distribution of neurons over these bins with an over-representation of neurons in the middle and higher SF and middle $n_x$. Importantly, the model completely lacks neurons with a combination of large SF and low $n_x$ and neurons with a combination of low SF and large $n_x$. Despite these differences in sampling, for the neuronal sub-populations present in both models, the response properties are similar and highly correlated (see Figure \ref{sup:act_spars_model_corr}).

\begin{figure} [h!]
\centering
\begin{tikzpicture}
\pgftext{\includegraphics[width=1\linewidth]{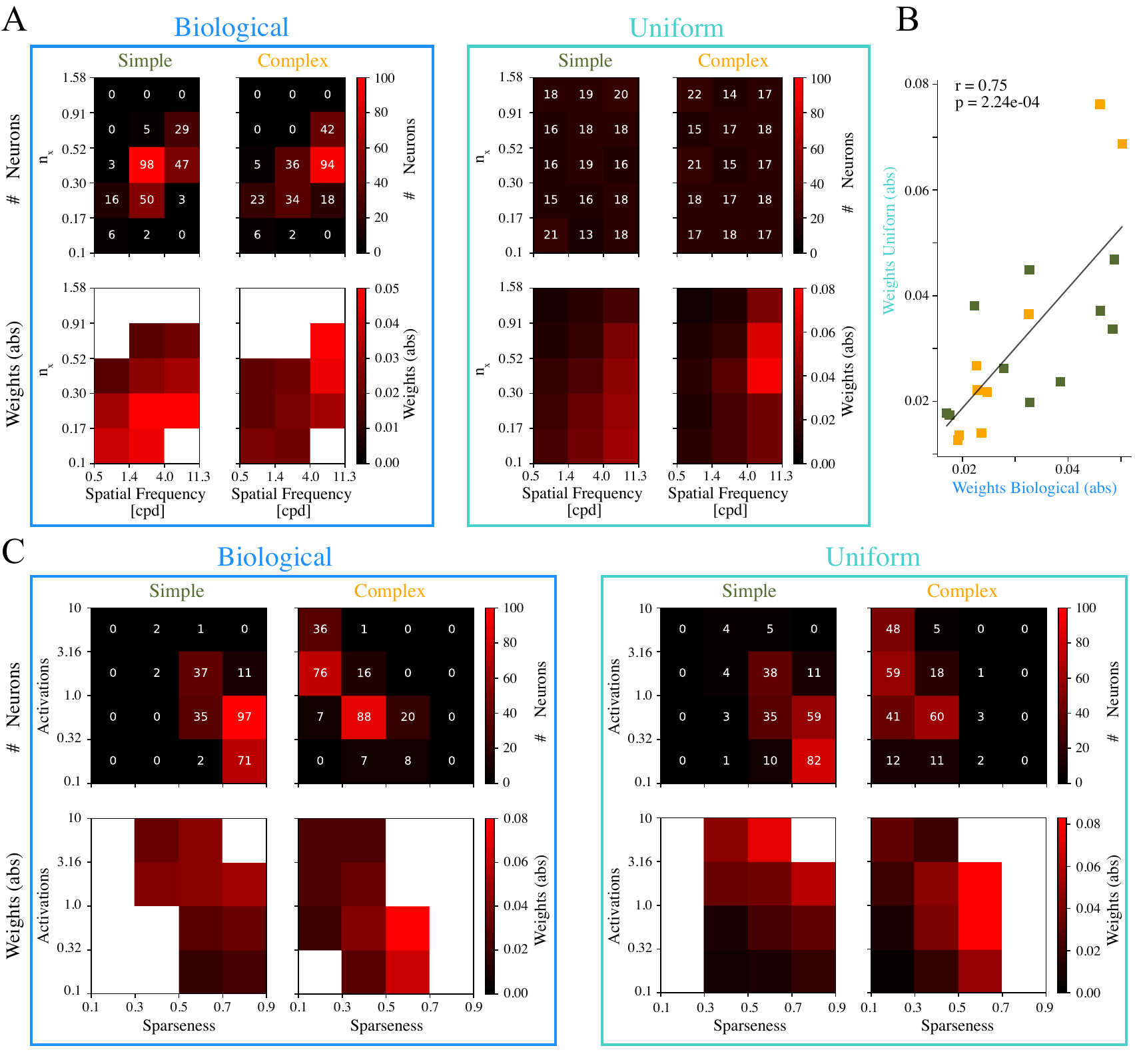}};
\end{tikzpicture}
\caption{\textbf{Downstream layer allocates similar weights to neuronal populations with similar RF properties in both model variants.} \textbf{A} Left, Biological Variant; right, Uniform Variant. Colormaps represent the average of all seeds ($n=4)$. Top row, distribution of number of neurons binned by SF and $n_x$ for simple and complex cells (each bin corresponds to neurons with similar GFB parameters). Bottom row, mean downstream absolute weights for neurons according to the same binning. Note that despite the differences in the neuronal sampling for the two models, the downstream absolute weights of neurons with similar properties are similar in both variants. \textbf{B} Mean downstream absolute weights per bin are correlated between the two model variants (r=0.75, p=2.24e-4, n=19 bins). Correlation is only calculated in the bins containing neurons in both models (ignores bins without neurons for the Biological Variant). \textbf{C} Similar to \textbf{A} but with neurons binned by mean response activation and sparseness (each bin corresponds to neurons with similar response properties).} \label{fig:4}
\end{figure}

How do the GFB parameters and the neuronal response properties affect the downstream weights of the V1 channels? To address this question, in the addition to grouping neurons by GFB parameters we also binned the neuronal sub-populations by response property (mean activation and sparseness) and calculated the mean absolute downstream weights after training the models (Figure \ref{fig:4}). For the neuronal sub-populations present in both models, the learned downstream absolute weights were highly correlated between model variants (Figure \ref{fig:3} A and B). Interestingly, despite the differences in sampling, there was an overlap between the distribution of neurons by response properties for the two variants, and their mean absolute downstream weights were also correlated (r=0.87, p=3.42e-6, 18 bins).

\begin{figure} [h!]
\centering
\begin{tikzpicture}
\pgftext{\includegraphics[width=0.9\linewidth]{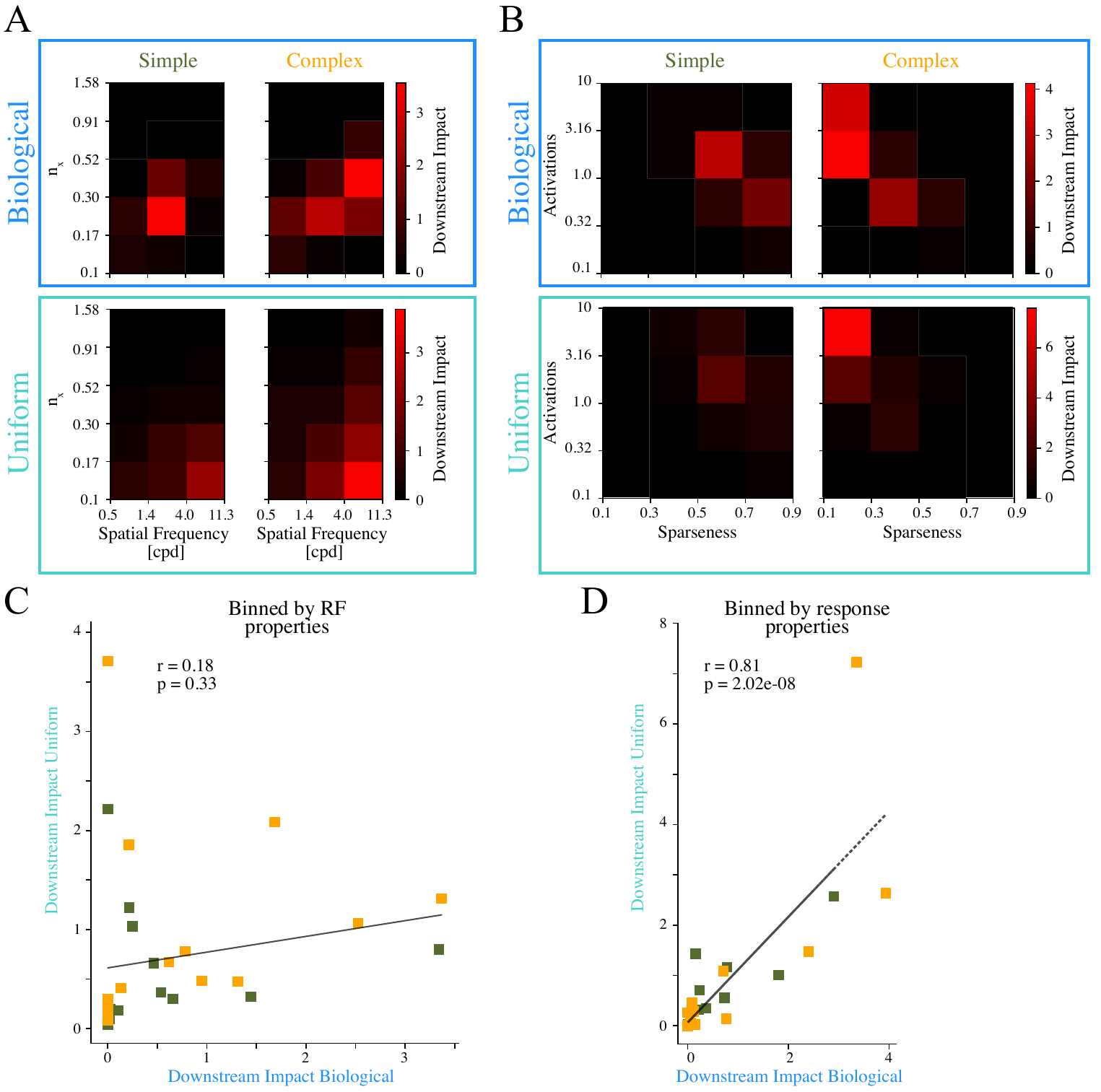}};
\end{tikzpicture}
\caption{\textbf{Impact on downstream layers from neurons with similar RF properties is different between model variants.} \textbf{A} Top, Biological Variant; down, Uniform Variant. Colormaps represent the average of all seeds ($n=4)$. Downstream impact of neurons binned by SF and $n_x$ for simple and complex cells (each bin corresponds to neurons with similar GFB parameters). Downstream Impact is calculated as the product of the number of neurons, the mean response activation, and the mean downstream absolute weights. Note that the bins that contribute the most to downstream layers are different for both model variants. \textbf{B} Similar to \textbf{A} but with neurons binned by mean response activation and sparseness (each bin corresponds to neurons with similar response properties). In this case the bins with  larger impact on downstream layers are similar between model variants. \textbf{C} Downstream impact per RF properties bin is not correlated between the two model variants (r=0.18, p=0.33, n=30 bins). Correlation is calculated in all the bins (bins without neurons have zero impact in downstream layers). \textbf{D} Downstream impact per response properties bin is correlated between the two model variants (r=0.81, p=2.02e-8, n=32 bins).} \label{fig:5}
\end{figure}

These analyses show a striking similarity between the two model variants. Despite the known difference in the sampling of the RF properties (GFB parameters), the response properties, as well as the learned absolute downstream weights are very similar between the two variants in the common neuronal sub-populations. We then estimated the impact on downstream layers for each neuronal sub-population. To obtain this quantity, we multiplied the number of channels in each sub-population by the mean response activation, and by the mean absolute downstream weight (Figure \ref{fig:5}). In the uniform variant, there is a disproportionate contribution of neurons with simultaneously high SF and high $n_x$. On the other hand, the model with biological sampling, this sub-population does not contribute at all, and instead most impact comes from neurons tuned to middle to high SF and with middle $n_x$. Downstream impact of the different neuronal sub-populations is therefore uncorrelated between the two variants (r=0.18, p=0.33, n=30 bins). Interestingly, when looking at the downstream impact grouped by the neuronal response properties, both variants show a more similar behavior (Figure \ref{fig:5} B and D).

\section{Discussion}
\label{sec:discussion}

The recent finding that adding a V1-like front-end to CNNs leads to improvements in robustness is of great importance as it suggests that biological intelligence can still contribute to improving machine learning models. However, understanding exactly where these improvements originate is often difficult to assess due to the complexity of these models.

In this work, we shed some light on which features lead to the improved robustness in a model with a V1-like front-end. To disambiguate the more general contribution of having V1-like RFs from precisely matching the population-level statistics of RF properties in primate V1 (V1 neuronal representations), we created two model variants of the VOneNet, a popular biologically-inspired CNN family \citep{dapello_simulating_2020, baidya_combining_2021, cirincione_implementing_nodate}. One of the variants sampled the RF properties from biological distributions (from available neurophysiological studies), precisely mimicking the neuronal representations found in primate V1, while the other variant sampled the RF properties uniformly within the same range. This difference in neuronal sampling was sufficient to explain the observed large difference in robustness to common image corruptions.

Interestingly, the response properties, as well as the learned weights in the common neuronal sub-populations were correlated between the two model variants. However, the non-uniform sampling of the biological distribution was critical to shift the neuronal sub-populations that had the strongest impact in downstream layers: from neurons with very large preferred SFs and very small RF sizes to neurons with larger RF sizes and tuned to somewhat smaller SFs.

One potential criticism of this result is that the model with biological sampling acts as a low-pass filter while the uniform model does not, helping it to deal with images corrupted with noise. However, this is not true as both models contain neurons with high SFs in similar numbers (Figure \ref{fig:1} C), and the fact that the biological model is marginally better in recognizing clean images. 

Importantly, when comparing both model variants, there is no trade-off between the uniform sampling and the biological sampling. Precisely matching the neuronal representations in primate V1 (biological sampling) leads to better robustness against image perturbations in all 15 types of corruptions tested. This results suggests that the neuronal representations in primate V1 are optimized to deal not only with clean images but also with a wide variety of different image distributions.





\acks{This work was developed within the scope of the project no. 62 - "Responsible AI", financed by European Funds, namely "Recovery and Resilience Plan - Component 5: Agendas Mobilizadoras para a Inovação Empresarial", included in the NextGenerationEU funding program.}

\bibliography{pmlr_paper}

\begin{thebibliography}{23}
\providecommand{\natexlab}[1]{#1}
\providecommand{\url}[1]{\texttt{#1}}
\expandafter\ifx\csname urlstyle\endcsname\relax
  \providecommand{\doi}[1]{doi: #1}\else
  \providecommand{\doi}{doi: \begingroup \urlstyle{rm}\Url}\fi

\bibitem[Adelson and Bergen(1985)]{Adelson1985}
E.H. Adelson and J.R. Bergen.
\newblock Spatiotemporal energy models for the perception of motion.
\newblock \emph{Journal of the Optical Society of America A, Optics and image
  science}, 2\penalty0 (2):\penalty0 284--299, 1985.
\newblock ISSN 1084-7529.
\newblock \doi{10.1364/JOSAA.2.000284}.
\newblock URL
  \url{http://www.opticsinfobase.org/abstract.cfm?URI=josaa-2-2-284}.
\newblock ISBN: 0740-3232 (Print).

\bibitem[Baidya et~al.(2021)Baidya, Dapello, DiCarlo, and
  Marques]{baidya_combining_2021}
Avinash Baidya, Joel Dapello, James~J. DiCarlo, and Tiago Marques.
\newblock Combining {Different} {V1} {Brain} {Model} {Variants} to {Improve}
  {Robustness} to {Image} {Corruptions} in {CNNs}.
\newblock \emph{Workshop on Shared Visual Representations in Human and Machine
  Intelligence (SVRHM 2021) of the Neural Information Processing Systems
  (NeurIPS)}, October 2021.
\newblock URL \url{http://arxiv.org/abs/2110.10645}.
\newblock arXiv: 2110.10645.

\bibitem[Cirincione et~al.(2022)Cirincione, Verrier, Bic, Olaiya, DiCarlo,
  Udeigwe, and Marques]{cirincione_implementing_nodate}
Andrew Cirincione, Reginald Verrier, Artiom Bic, Stephanie Olaiya, James~J
  DiCarlo, Lawrence Udeigwe, and Tiago Marques.
\newblock Implementing {Divisive} {Normalization} in {CNNs} {Improves}
  {Robustness} to {Common} {Image} {Corruptions}.
\newblock \emph{Workshop on Shared Visual Representations in Human and Machine
  Intelligence (SVRHM 2022) of the Neural Information Processing Systems
  (NeurIPS)}, 2022.

\bibitem[Dapello et~al.(2020)Dapello, Marques, Schrimpf, Geiger, Cox, and
  DiCarlo]{dapello_simulating_2020}
Joel Dapello, Tiago Marques, Martin Schrimpf, Franziska Geiger, David~D. Cox,
  and James~J. DiCarlo.
\newblock Simulating a primary visual cortex at the front of {CNNs} improves
  robustness to image perturbations.
\newblock \emph{NeurIPS}, pages 1--30, 2020.
\newblock ISSN 26928205.
\newblock \doi{10.1101/2020.06.16.154542}.

\bibitem[De~Valois et~al.(1982{\natexlab{a}})De~Valois, Albrecht, and
  Thorell]{DeValois1982531}
Russell~L. De~Valois, Duane~G. Albrecht, and Lisa~G. Thorell.
\newblock Spatial {Frequency} {Selectivity} of {Cells} in {Macaque} {Visual}
  {Cortex}.
\newblock \emph{Vision Research}, 22:\penalty0 545--559, 1982{\natexlab{a}}.

\bibitem[De~Valois et~al.(1982{\natexlab{b}})De~Valois, Yund, and
  Hepler]{DeValois1982545}
Russell~L. De~Valois, E.~W. Yund, and Norva Hepler.
\newblock The orientation and direction selectivity of cells in macaque visual
  cortex.
\newblock \emph{Vision Research}, 22:\penalty0 531--544, 1982{\natexlab{b}}.

\bibitem[Dodge and Karam(2017)]{dodge_study_2017}
Samuel Dodge and Lina Karam.
\newblock A {Study} and {Comparison} of {Human} and {Deep} {Learning}
  {Recognition} {Performance} {Under} {Visual} {Distortions}.
\newblock \emph{arXiv:1705.02498 [cs]}, May 2017.
\newblock URL \url{http://arxiv.org/abs/1705.02498}.
\newblock arXiv: 1705.02498.

\bibitem[Evans et~al.(2022)Evans, Malhotra, and Bowers]{EVANS202296}
Benjamin~D. Evans, Gaurav Malhotra, and Jeffrey~S. Bowers.
\newblock Biological convolutions improve dnn robustness to noise and
  generalisation.
\newblock \emph{Neural Networks}, 148:\penalty0 96--110, 2022.
\newblock ISSN 0893-6080.
\newblock \doi{https://doi.org/10.1016/j.neunet.2021.12.005}.
\newblock URL
  \url{https://www.sciencedirect.com/science/article/pii/S0893608021004780}.

\bibitem[Geirhos et~al.(2018)Geirhos, Temme, Rauber, Schütt, Bethge, and
  Wichmann]{geirhos_generalisation_2018}
Robert Geirhos, Carlos R.~Medina Temme, Jonas Rauber, Heiko~H. Schütt,
  Matthias Bethge, and Felix~A. Wichmann.
\newblock Generalisation in humans and deep neural networks.
\newblock In \emph{{NeurIPS}}, pages 1--13, 2018.
\newblock URL \url{http://arxiv.org/abs/1808.08750}.

\bibitem[He et~al.(2016)He, Zhang, Ren, and Sun]{he_deep_2016}
Kaiming He, Xiangyu Zhang, Shaoqing Ren, and Jian Sun.
\newblock Deep {Residual} {Learning} for {Image} {Recognition}.
\newblock In \emph{{CVPR}}, pages 1--12, December 2016.
\newblock URL \url{http://arxiv.org/abs/1512.03385}.

\bibitem[Hendrycks and Dietterich(2019)]{hendrycks_benchmarking_2019}
Dan Hendrycks and Thomas Dietterich.
\newblock Benchmarking neural network robustness to common corruptions and
  perturbations.
\newblock \emph{7th International Conference on Learning Representations, ICLR
  2019}, pages 1--16, 2019.

\bibitem[Jones and Palmer(1987)]{jones_two-dimensional_1987}
J.~P. Jones and L.~A. Palmer.
\newblock The two-dimensional spatial structure of simple receptive fields in
  cat striate cortex.
\newblock \emph{Journal of Neurophysiology}, 58\penalty0 (6):\penalty0
  1187--1211, 1987.
\newblock ISSN 00223077.
\newblock \doi{10.1152/jn.1987.58.6.1187}.

\bibitem[Krizhevsky et~al.(2012)Krizhevsky, Sutskever, and
  Geoffrey~E.]{Krizhevsky2012}
Alex Krizhevsky, Ilya Sutskever, and Hinton Geoffrey~E.
\newblock {ImageNet} {Classification} with {Deep} {Convolutional} {Neural}
  {Networks}.
\newblock In \emph{{NIPS}}, pages 1097--1105, 2012.
\newblock ISBN 978-1-62748-003-1.
\newblock \doi{10.1109/5.726791}.
\newblock ISSN: 10495258.

\bibitem[Le and Yang(2015)]{le_tiny_nodate}
Ya~Le and Xuan Yang.
\newblock Tiny {ImageNet} {Visual} {Recognition} {Challenge}.
\newblock pages 1--6, 2015.

\bibitem[Malhotra et~al.(2020)Malhotra, Evans, and
  Bowers]{malhotra_hiding_2020}
Gaurav Malhotra, Benjamin~D. Evans, and Jeffrey~S. Bowers.
\newblock Hiding a plane with a pixel: examining shape-bias in {CNNs} and the
  benefit of building in biological constraints.
\newblock \emph{Vision Research}, 174:\penalty0 57--68, September 2020.
\newblock ISSN 00426989.
\newblock \doi{10.1016/j.visres.2020.04.013}.
\newblock URL
  \url{https://linkinghub.elsevier.com/retrieve/pii/S0042698920300742}.

\bibitem[Ringach(2002)]{Ringach2002b}
Dario~L Ringach.
\newblock Spatial {Structure} and {Symmetry} of {Simple}-{Cell} {Receptive}
  {Fields} in {Macaque} {Primary} {Visual} {Cortex}.
\newblock \emph{Journal of Neurophysiology}, \penalty0 (88):\penalty0 455--463,
  2002.

\bibitem[Rust et~al.(2005)Rust, Schwartz, Movshon, and Simoncelli]{Rust2005}
Nicole~C Rust, Odelia Schwartz, J.~A. Movshon, and E.~P. Simoncelli.
\newblock Spatiotemporal {Elements} of {Macaque} {V1} {Receptive} {Fields}.
\newblock \emph{Neuron}, 46:\penalty0 945--956, 2005.
\newblock \doi{10.1016/j.neuron.2005.05.021}.

\bibitem[Schiller et~al.(1976)Schiller, Finlay, and Volman]{Schiller1976}
P.~H. Schiller, B.~L. Finlay, and S.~F. Volman.
\newblock Quantitative studies of single-cell properties in monkey striate
  cortex. {III}. {Spatial} {Frequency}.
\newblock \emph{Journal of neurophysiology}, 39\penalty0 (6):\penalty0
  1334--1351, 1976.
\newblock ISSN 0022-3077.
\newblock \doi{10.1152/jn.1976.39.6.1352}.
\newblock URL \url{http://www.ncbi.nlm.nih.gov/pubmed/825624}.

\bibitem[Simonyan and Zisserman(2015)]{simonyan_very_2015}
Karen Simonyan and Andrew Zisserman.
\newblock Very {Deep} {Convolutional} {Networks} for {Large}-{Scale} {Image}
  {Recognition}.
\newblock In \emph{{ICLR}}, pages 1--14, 2015.
\newblock ISBN 1097-0142 (Electronic){\textbackslash}n0008-543X (Linking).
\newblock \doi{10.2146/ajhp170251}.
\newblock URL \url{http://arxiv.org/abs/1409.1556}.
\newblock ISSN: 15352900.

\bibitem[Softky and Koch(1993)]{Softky1993}
W.~R. Softky and C.~Koch.
\newblock The highly irregular firing of cortical cells is inconsistent with
  temporal integration of random {EPSPs}.
\newblock \emph{Journal of Neuroscience}, 13\penalty0 (1):\penalty0 334--350,
  1993.
\newblock ISSN 02706474.
\newblock \doi{10.1523/jneurosci.13-01-00334.1993}.

\bibitem[Szegedy et~al.(2015)Szegedy, Liu, Jia, Sermanet, Reed, Anguelov,
  Erhan, Vanhoucke, and Rabinovich]{szegedy_going_2015}
Christian Szegedy, Wei Liu, Yangqing Jia, Pierre Sermanet, Scott Reed, Dragomir
  Anguelov, Dumitru Erhan, Vincent Vanhoucke, and Andrew Rabinovich.
\newblock Going deeper with convolutions.
\newblock \emph{Proceedings of the IEEE Computer Society Conference on Computer
  Vision and Pattern Recognition}, 07-12-June:\penalty0 1--9, 2015.
\newblock ISSN 10636919.
\newblock \doi{10.1109/CVPR.2015.7298594}.
\newblock ISBN: 9781467369640.

\bibitem[Vinje and Gallant(2000)]{vinje2000sparse}
William~E Vinje and Jack~L Gallant.
\newblock Sparse coding and decorrelation in primary visual cortex during
  natural vision.
\newblock \emph{Science}, 287\penalty0 (5456):\penalty0 1273--1276, 2000.

\bibitem[Virtanen et~al.(2020)Virtanen, Gommers, Oliphant, Haberland, Reddy,
  Cournapeau, Burovski, Peterson, Weckesser, Bright, et~al.]{virtanen2020scipy}
Pauli Virtanen, Ralf Gommers, Travis~E Oliphant, Matt Haberland, Tyler Reddy,
  David Cournapeau, Evgeni Burovski, Pearu Peterson, Warren Weckesser, Jonathan
  Bright, et~al.
\newblock Scipy 1.0: fundamental algorithms for scientific computing in python.
\newblock \emph{Nature methods}, 17\penalty0 (3):\penalty0 261--272, 2020.

\end{thebibliography}

\newpage
\appendix

\section{Statistics}\label{apd:statistics}
For the correlations analysis, we defined statistical significance for p-value $< 0.05$. The exact p-values and r-values are indicated in each reported correlation. The r-values correspond to Pearson correlation coefficients. Correlations between the Biological and Uniform variants for responses and weights are only calculated in the bins containing neurons in both models (bins without neurons are ignored), whereas for downstream impact the correlation is calculated in all bins (bins without neurons have zero effect in downstream layers).
The statistical analyses were performed with the Scipy library for Python \cite{virtanen2020scipy}. 

\section{Datasets}\label{apd:datasets}
\subsection{Tiny ImageNet}
\label{sec:tiny}

We trained and evaluated models' clean accuracy on the Tiny ImageNet dataset \cite{le_tiny_nodate}. Tiny ImageNet contains 100.000 images of 200 classes (500 for each class) downsized to 64$\times$64 colored images. Each class has 500 training images, 50 validation images and 50 test images. The Tiny ImageNet dataset is publicly available at \url{https://www.kaggle.com/c/tiny-imagenet}.

\subsection{Tiny ImageNet-C (Common Corruptions)}
\label{sec:corruptions}

We used Tiny ImageNet-C \cite{hendrycks_benchmarking_2019} to evaluate model robustness to common corruptions. The Tiny ImageNet-C dataset consists of 15 different corruption types applied to validation images of Tiny ImageNet. The individual corruption types are grouped into 4 main categories: Noise (Gaussian noise, shot noise, impulse noise), Blur (defocus blur, glass blur, motion blur, zoom blur), Weather (snow, frost, fog, brightness) and Digital effects (contrast, elastic transform, pixelate and JPEG compression). Each of the 15 corruption types has 5 levels of severity, resulting in a total of 75 perturbations. The Tiny ImageNet-C is publicly available at \url{https://github.com/hendrycks/robustness} under Creative Commons Attribution 4.0 International.

\section{Models}
\label{sec:models}

\subsection{VOneNets}
\label{sec:models_vonenets}

\textsc{\bf VOneNet Model Family}
VOneNets \cite{dapello_simulating_2020} are hybrid CNNs, with a biologically-constrained fixed-weight front-end layer that simulates V1, called the VOneBlock, followed by a neural network back-end adapted from current CNN vision models. The VOneBlock is a linear-nonlinear-Poisson (LNP) model of V1 \cite{Rust2005}, consisting of a fixed-weight Gabor Filter Bank (GFB) \cite{jones_two-dimensional_1987}, simple and complex cell \cite{Adelson1985} nonlinearities, and neuronal stochasticity \cite{Softky1993}. The code for the VOneNet model family is publicly available at \url{https://github.com/dicarlolab/vonenet} under GNU General Public License v3.0.


\textsc{\bf Adapting VOneNets to Tiny ImageNet} 
To create the VOneNets, ResNet18 \cite{he_deep_2016} was chosen as the back-end architecture. We built the VOneResNet18 by removing the first block of ResNet18 (one stack of convolution, normalization, non-linearity and pooling layers) and replacing it with the VOneBlock and a trainable bottleneck layer (a transition layer used to compress the 512 channels to 64 which is the depth of the next layer of the back-end architecture). 
Since the standard VOneNets were adjusted to be used with ImageNet (224px input size), we made several modifications to adapt the VOneNet architecture to the Tiny ImageNet image size (64px), as in \cite{baidya_combining_2021}.  
To prevent the VOneBlock output from having a very small spatial map, the stride of the convolution layer (GFB) was set to 2 instead of 4. We also changed the input field of view from 8deg (for ImageNet) to 2deg (for Tiny ImageNet) to account for the fact that now images represent a narrower portion of the visual space. This change resulted in an input resolution -- number of pixels per degree (ppd) -- of 32 ppd for Tiny ImageNet, similar to that of ImageNet (28 ppd). 


\textsc{\bf VOneResNet18 Variants} 
We created two VOneResNet18 model variants by modifying the GFB parameters sampling. For the VOneResNet18 Biological model, the GFB parameters are randomly sampled from empirically observed distributions of preferred orientation, peak spatial frequency (SF), and size/shape of receptive fields \cite{DeValois1982531, DeValois1982545, Ringach2002b}, whereas the in the VOneResNet18 Uniform variant the GFB parameters are generated by randomly sampling from uniform distributions.
In both variants, the channels are divided equally between simple- and complex-cells (256 each) and the SF is between $[0.5-11.3]$ cpd. 
To facilitate the comparison between the uniform and biological models, we removed the stochasticity generator, so that the models are noise-free.

\subsection{ResNet18}
\label{sec:models_resnet18}

We used a variant of the Torchvision implementation of ResNet18 \cite{he_deep_2016} as the base model and as the model back-end for both VOneResNet18 variants. The first block of the original ResNet18 model (the block replaced by VOneBlock in VOneResNet18) has a combined stride of 4 (2 in the convolution layer and 2 in the maxpool layer). To maintain the size of ResNet18 comparable to VOneResNet18, we adjusted the ResNet18 architecture so that it has a stride of 1 in the first convolutional layer and kept the stride of 2 in the maxpool layer, resulting in a combined stride of 2 in the first block which is the same as the VOneBlock. 

\subsection{Training}
\label{sec:models_training}

We used PyTorch version 1.10.2. All models were trained on a machine with 1$\times$32GB V100 GPU. For each of the three variants we trained four models, with four different seeds of the random generator.
The training procedure is detailed below.

\textsc{\bf Preprocessing} During training, preprocessing included scaling the images with a factor randomly sampled between 1-1.2, rotating the images with a rotation angle randomly sampled between -30 to 30 degrees, flipping the images horizontally with a random probability of 0.5, and shifting the images in the horizontal and vertical directions by a pixel distance randomly sampled between -5\% to 5\% of the image width and height, respectively. Images were normalized by subtraction and division by [0.5, 0.5, 0.5]. 

During evaluation, preprocessing only involved image normalization, i.e. subtraction and division by [0.5, 0.5, 0.5].

\textsc{\bf Loss Functions} The loss function was given by the cross-entropy loss between image labels and model predictions (logits).  

\textsc{\bf Optimization} We used Stochastic Gradient Descent with an initial learning rate 0.1, momentum 0.9, and a weight decay of 0.0005. The learning rate was dynamically adjusted by dividing it by 10 whenever there was no significant improvement (threshold of 0.01) in validation loss for 5 consecutive epochs. All models were trained using a batch size of 128 images for 60 epochs.

\section{Detailed Accuracies} 
\label{sec:detailed_accuracies}

All reported accuracies represent the top-1 accuracy on the validations datasets of Tiny ImageNet (for clean) and Tiny ImageNet-C (for corruptions). In the main text and Figure \ref{fig:1} D the accuracies are reported relative to the standard model ResNet18 and in the Appendices (Table 1 and Figure \ref{sup:detailed_accuracies}) they are absolute values. 

\begin{table} [h!] 
\label{tab:absolute_accuracies}%
 \caption{\textbf{Absolute accuracies of ResNet18, VOneResNet18 Biological and VOneResNet18 Uniform}. Clean images and 15 types of common image corruptions (averaged over five perturbation severities). The value in parenthesis represents the standard error of the mean (n = 4 seeds). The VOneResNet18 Biological has the highest accuracy in all image types except clean, contrast, and pixelate.}  
  \small
 \centering
 \begin{tabular}{@{}l@{\hskip 10pt}c@{\hskip 10pt}ccc@{\hskip 10pt}cccc} \toprule
 & & \multicolumn{3}{c}{Noise}{\hskip 10pt} & \multicolumn{4}{c}{Blur} \\ \cmidrule(r{10pt}){3-5} \cmidrule(r){6-9}
 & Clean & Gaussian & Shot & Impulse & Defocus & Glass & Motion & Zoom \\
Model & [\%] & [\%] & [\%] & [\%] & [\%] & [\%] & [\%]  & [\%] \\ \midrule
ResNet18 & \textbf{57.7} & 19.4 & 22.6 & 21.4 & 14.2 & 19.0 & 19.5 & 16.2 \\ 
& (0.27) & (0.19) & (0.26) & (0.16) & (0.32) & (0.21) & (0.49) & (0.57) \\
VOneResNet18 Uniform & 55.7 & 20.2 & 24.2 & 21.5 & 13.5 & 18.7 & 18.7 & 15.3 \\ 
& (0.32) & (0.51) & (0.6) & ( 0.25) & (0.56) & (0.25) & (0.6) &  (0.59) \\
VOneResNet18 Biological & 56.6 & \textbf{22.8} & \textbf{27.2} & \textbf{22.6} & \textbf{14.7} & \textbf{19.1} & \textbf{20.3} & \textbf{16.5} \\ 
& (0.25) & (0.23) & (0.33) & (0.22) & (0.32) & (0.2) &  (0.3) &  (0.24) \\
\bottomrule
\quad
\end{tabular}
 \begin{tabular}{@{}l@{\hskip 10pt}ccccc@{\hskip 10pt}cccc} \toprule
 & \multicolumn{4}{c}{Weather}{\hskip 10pt} & \multicolumn{4}{c}{Digital} \\ \cmidrule(r{10pt}){2-5} \cmidrule(r){6-9}
 & Snow & Frost & Fog & Bright. & Contrast & Elastic & Pixelate & JPEG \\
Model & [\%] & [\%] & [\%] & [\%] & [\%] & [\%] & [\%] & [\%] \\ \midrule
ResNet18 & 23.2 & 24.7 & 21.2 & 27.1 & \textbf{9.7} & 24.5 & \textbf{37.8} & 31.7 \\ 
& (0.57) & (0.27) & (0.29) & (0.25) & (0.14) & (0.45) & (0.35) & (0.41) \\
VOneResNet18 Uniform & 23.9 & 24.1 & 21.1 & 25.7 & 8.5 & 24.4 & 34.8 & 32.6 \\ 
& (0.76) & (0.73) & (0.4) &  (0.52) & (0.19) & (0.74) & (0.4) &  (0.87) \\
VOneResNet18 Biological & \textbf{27.3} & \textbf{27.2} & \textbf{22.4} & \textbf{28.8} & 9.3 & \textbf{26.9} & 36.5 & \textbf{36.8} \\ 
& (0.49) & (0.5) &  (0.33) & (0.42) & (0.19) & (0.42) & (0.11) & (0.42) \\
\bottomrule
\end{tabular}
\end{table}

\begin{figure} [h!]
\centering
\begin{tikzpicture}
\pgftext{\includegraphics[width=0.6\linewidth]{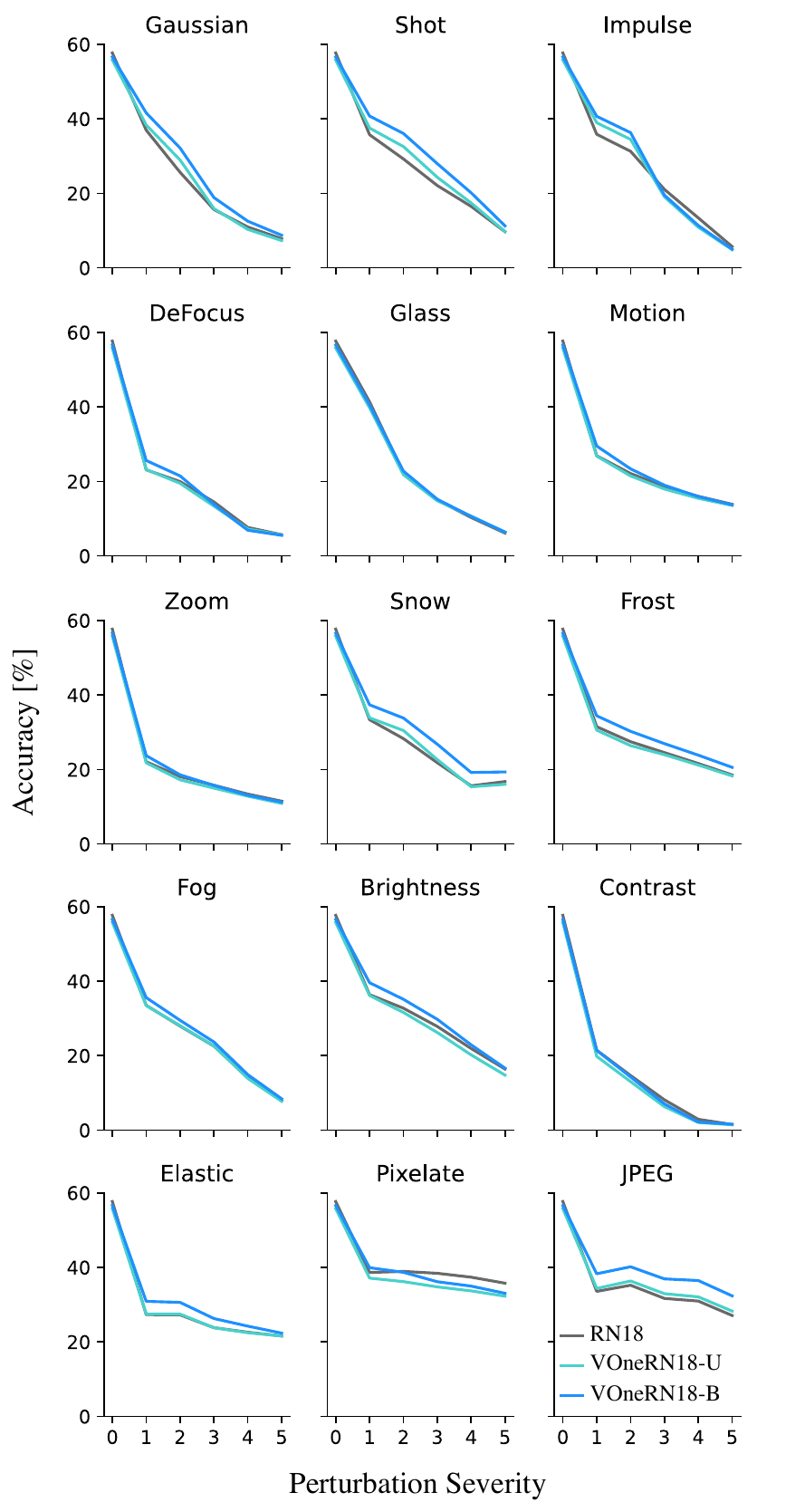}};
\end{tikzpicture}
\caption{\textbf{VOneResNet18 Biological shows consistent improvements in robustness.} Absolute accuracies for ResNet18 (RN18, gray), VOneResNet18 Biological (VOneRN18-B, blue), VOneResNet18 Uniform (VOneRN18-U, turquoise) for the 15 corruption types at all perturbation severity levels. Accuracy represents top-1 (\%). Perturbation 0 corresponds to no corruption (clean images).} \label{sup:detailed_accuracies}
\end{figure}

\newpage
\section{Neuronal Representations Analyses} 
\label{sec:representation_analyses}

\subsection{GFB parameters}
\label{sec:gfb_parameters}

The GFB contains Gabor filters with of multiple orientations, spatial frequencies, and sizes.
The parameters are randomly sampled from distributions of preferred orientation ($[0-180]$ deg), peak spatial frequency ($[0.5-11.3]$ cpd), and size/shape of receptive fields (both $n_x$ and $n_y$ in $[0.1-1.585]$). The Biological variant samples the GFB parameters from empirical distributions \citep{DeValois1982531, DeValois1982545, Ringach2002b, Schiller1976}. While orientation is sampled independently, the RF sizes along the perpendicular and parallel axes are sampled from a joint distribution, and a correlation is introduced between the RF size and the SF to match what is observed empirically (Figure \ref{fig:1}).

\begin{figure} [h!]
\centering
\begin{tikzpicture}
\pgftext{\includegraphics[width=0.6\linewidth]{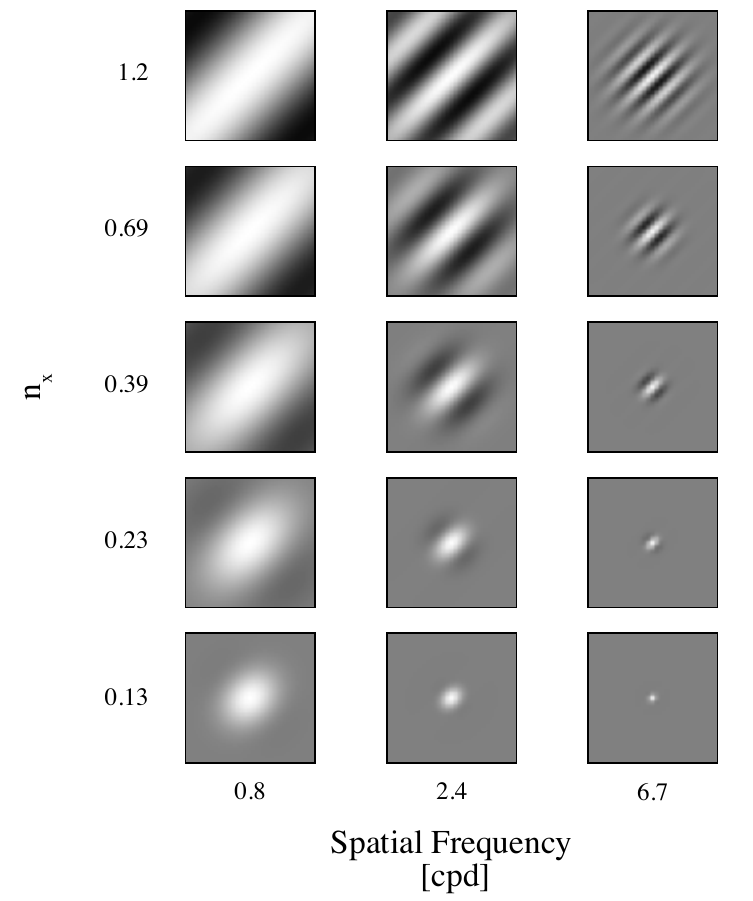}};
\end{tikzpicture}
\caption{\textbf{RFs of neurons with different Spatial Frequency and $n_x$.} Visualization of the  RF kernels for the neurons in the center of each Spatial Frequency and $n_x$ bins as shown in Figures \ref{fig:3}, \ref{fig:4}, and \ref{fig:5}.} \label{sup:gabor}
\end{figure}

Figure \ref{sup:gabor} shows Gabor filters with a preferred orientation of 45 degrees, different spatial frequencies and $n_x$ (values are chosen in the center of the bins for each parameter).

\subsection{Response properties}
\label{sec:rf_properties}

We studied the relationship between response properties (mean activation, and sparseness) and the neuronal RF properties (SF, size, orientation). The mean activation for a channel was calculated as the average over the neurons and over a batch of 1000 images.
The sparseness was calculated according to the formula in \cite{vinje2000sparse} as follows: for a specific channel and a specific neuron, $S=\frac{1-\left(\left(\sum_{k=1}^b{(a_k)}/b\right)^2/\sum_{k=1}^b{(a_k^2/b)}\right)}{1-1/b}$, where $b$ is the batch size (here 1000 images) and $a_k$ is the response activation to the $k$th image in the batch. Then the mean sparseness for a channel was calculated as the mean of $S$ over the neurons.

\begin{figure} [h!]
\centering
\begin{tikzpicture}
\pgftext{\includegraphics[width=0.75\linewidth]{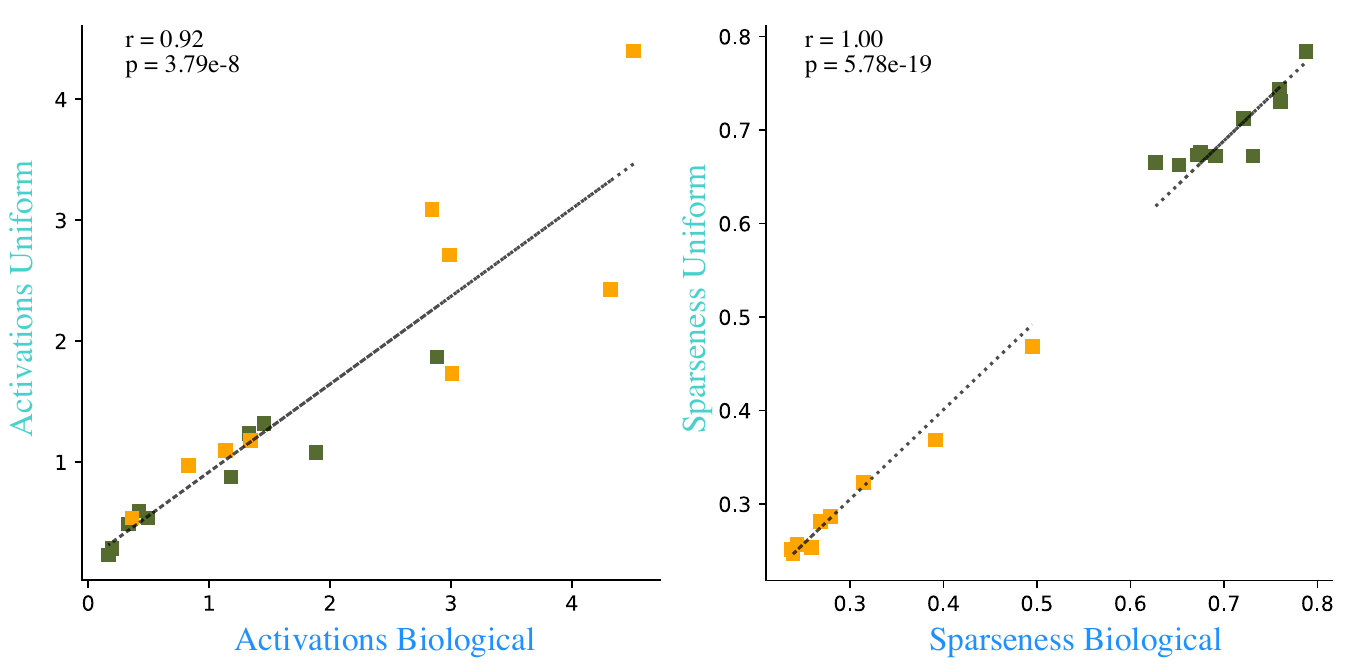}};
\end{tikzpicture}
\caption{\textbf{Response activation and sparseness of different neuronal sub-populations are correlated across model variants.} Left, mean response activations of neuronal sub-populations (binned by RF properties) for the two VOneNetResNet18 variants. Green squares represent simple cells and yellow squares represent complex cells. Right, same as in left but for response sparseness. Relative to Fig. \ref{fig:3}.} 
\label{sup:act_spars_model_corr}
\end{figure}

\subsection{Downstream Weights}
\label{sec:weights}

To get an understanding of the importance the models place on different channels, we examined the weights of the bottleneck layer that follows the VOneBlock in the VOneResNet18 models. The mean absolute downstream weights are calculated by averaging the absolute weights of the bottleneck layer over its 64 outputs.

\subsection{Downstream Impact}
\label{sec:impact}

The Downstream Impact for a population of neurons was defined as the effect of those neurons in the downstream layers of the network. We evaluated the Downstream Impact of a population of neurons by multiplying the number of neurons in the population with the mean response activations of the neurons in the VOneBlock and with the mean absolute weights of the bottleneck layer. The natural consequence of this definition is that empty populations will have no impact in downstream layers.

\end{document}